# Modeling and Optimization of Epidemiological Control Policies Through Reinforcement Learning

Ishir Rao[1]

1. Chatham High School, Chatham, New Jersey

# 1. ABSTRACT

Pandemics involve the high transmission of a disease that impacts global and local health and economic patterns. The impact of a pandemic can be minimized by enforcing certain restrictions on a community. However, while minimizing infection and death rates, these restrictions can also lead to economic crises. Epidemiological models help propose pandemic control strategies based on non-pharmaceutical interventions such as social distancing, curfews, and lockdowns, reducing the economic impact of these restrictions. However, designing manual control strategies while considering disease spread and economic status is non-trivial. Optimal strategies can be designed through multi-objective reinforcement learning (MORL) models, which demonstrate how restrictions can be used to optimize the outcome of a pandemic. In this research, we utilized an epidemiological Susceptible, Exposed, Infected, Recovered, Deceased (SEIRD) model – a compartmental model for virtually simulating a pandemic day by day. We combined the SEIRD model with a deep double recurrent Q-network to train a reinforcement learning agent to enforce the optimal restriction on the SEIRD simulation based on a reward function. We tested two agents with unique reward functions and pandemic goals to obtain two strategies. The first agent placed long lockdowns to reduce the initial spread of the disease, followed by cyclical and shorter lockdowns to mitigate the resurgence of the disease. The second agent provided similar infection rates but an improved economy by implementing a 10-day lockdown and 20-day no-restriction cycle. This use of reinforcement learning and epidemiological modeling allowed for both economic and infection mitigation in multiple pandemic scenarios.

# 2. INTRODUCTION

During a pandemic, the foremost intention is to minimize any loss, such as deaths and economic crisis, caused by the disease. This can be completed through proper medical treatment and reducing disease spread. However, medical resources are not always sufficient to treat a large infected population, as is common in pandemics. Consequently, effectively reducing disease spread is vital to minimize the loss caused by a pandemic situation (1).

To control the outcome of pandemics such as the COVID-19 pandemic caused by the SARS-CoV-2 coronavirus, we must ensure that any infection is spread gradually to prevent medical resources from becoming overwhelmed by the medical attention required. Moreover, we must control any resurgence of the disease. Another factor that we must take into consideration is the economic status of the population. By optimizing the economy, we can ensure that the population can return to a state of normalcy post-pandemic, rather than an economic crisis (2).

This task of minimizing a factor (the spread of disease) while maximizing another (the economic status) is known as an optimization problem. We utilized reinforcement learning (RL), a subsection of machine learning, to gather the knowledge for the best decisions in the optimization problem, which is referred to as the policy. An RL agent can interact with an environment through distinct actions and is given a corresponding reward for each action (3). By looping through this, the RL agent can formulate the optimal policy which maximizes the reward, demonstrating that the agent can learn similarly to an intelligent being. The agent, the environment, and the rewards can be thought of as an escape room, where the player ("agent") receives rewards and clues as they progress and "maximize" how well they are doing in the room environment while minimizing the cost.

Previously, RL was limited by functions such as Q-Learning and Double Q-Learning, which are tabular in nature. These functions only result in one output for each input value and generate a function over time (3). However, with the recent implementation of deep neural networks, deep reinforcement learning (DRL) techniques such as Deep Q-Learning and Deep Double Q-Learning have proven to be highly accurate and converge to the optimal policy faster than traditional techniques (4-6). This is completed by using a neural network to store the input-output information and using trends in the data to then predict the results of future unknown inputs.

In this work, we implemented a short-term memory-based Deep Double Q-Network (DDQN) to find the optimal policy for pandemic control (5). A DDQN is used over a Deep Q-Network (DQN) since they are able to use two separate networks to decouple action selection from value estimation, whereas DQNs do this in one network. By separating action selection and value estimation, DDQNs are able to reduce overestimation biases and converge to the optimal policy faster. We utilized a widely used epidemiological model known as the Susceptible, Exposed, Infected, Recovered, Deceased (SEIRD) model, which provided us with a simulated environment of a pandemic (7). The DDQN agent explored this environment and acted as the "government" by implementing restrictive measures (such as social distancing and lockdowns) to find a policy that optimized the outcome of the pandemic and minimizing economic downturn. These pandemic control strategies could then be used and implemented by a nation's government. As a result, it is important to note that, for the simulation to match the real-world environment, regulations would have to be enforced by local, state, and national forces, and guidelines would have to be followed perfectly by people, which may not necessarily happen.

This use of RL and epidemiological modeling allowed for both economic and infection mitigation in multiple pandemic scenarios. After testing for two unique scenarios, we found that these methods optimized pandemic outcomes by minimizing the economic impact and spread of infection.

## 3. RESULTS

We utilized the SEIRD model to test four levels of restriction on an environment: no restriction (control), social distancing, lockdown, and a lockdown with a curfew (Figure 1). In these four single-action simulations, no reinforcement learning agents were used. All tests were given identical initial conditions for the environment (SEIRD model), and these same initial conditions are later used when the agents operated on the environment. Each of the SEIRD simulations in this work ends when the number of infected individuals reaches 0, indicating that the disease has been eradicated. We also utilized an economic model to track the state of the economy throughout a pandemic simulation. We measure the economic state on each day of the pandemic, and later graph this alongside the SEIRD model. It is important to note that, in this research, we simplified the "economy" to be the number of people in the workforce, rather than a nuanced economic model with numerous factors, which made it more efficient for this research. The simplified economic model sufficiently provided a numerical value for the economy, which the RL agent used to learn, and this value is an abstract estimate of the state of the economy and does not use any specific units. An RL agent also uses a reward function that uses current environmental factors to calculate a reward value that represents how well the agent is accomplishing the assigned goal. The reward function is used by the agent to determine which actions reduce or increase the reward and learn from the results, and the reward value is an arbitrary value which serves only to inform the agent on how it can improve.

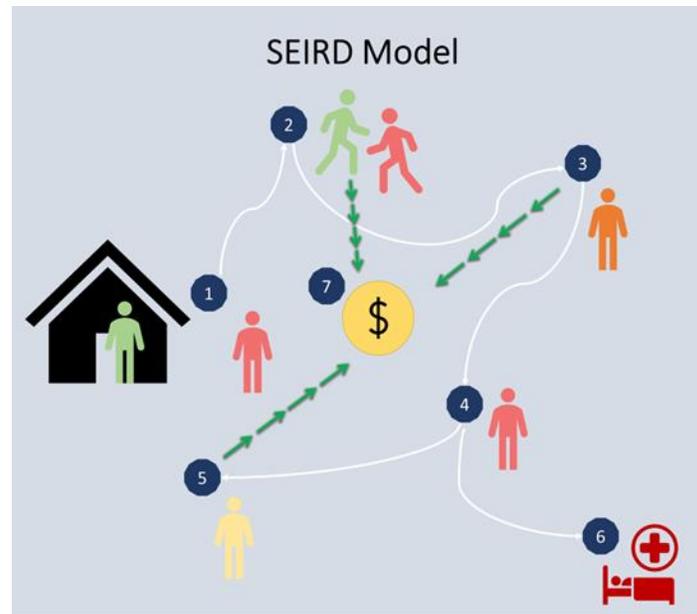

Figure 1. SEIRD compartmental model infographic. The different stages in the SEIRD model along with the economic component, where 1: Two groups begin separate – red is infected, green is susceptible. 2: A susceptible person comes in contact with an infected individual. Susceptible turns into exposed. 3: The exposed individual is a host of the disease but is not aware and cannot spread it yet. 4: Eventually, they can become fully infected. 5: After some time, proper medical care or self-immunity leads them to recover. 6: However, if medical care is not sufficient, they may not survive. 7: Susceptible, exposed, and recovered individuals contribute to the economy throughout. The SEIRD model was simulated computationally using solve_ivp from the SciPy module along with Matplotlib in Python. The infectious population began as 7% of the total population of 1000 individuals. The model was rerun for each training step to provide data to the RL agent.

All test graphs showed each of the SEIRD populations and the economic status based on each restrictive measure (Figure 2). Each test graph had a similar shape, indicating consistently inverse relationships between infected individuals vs. economic status, and susceptible vs. recovered individuals. It was seen that progressively stricter restrictions flattened the infection curve and managed infection and mortality rates more effectively. Along with this, the economic status (or normalcy) of the population reduced as restrictions became stricter. With no restriction, it took only 50 days for a quarter of the population to become infected, at which point the economy was at 60% normalcy (Figure 2a). On the other hand, with both a lockdown and curfew, the number of infections remained the same but the pandemic lasted much longer, taking 300 days for a quarter of the population to be infected, at which point the economy was at 15% normalcy (Figure 2d). This longer period of time before a quarter infection may allow for significantly more vaccine discovery and testing and, as a result, may be less likely to reach the widespread infection seen with no restrictions. Since the SEIRD model is unable to predict and consider vaccinations for diseases, this was not a factor within our predictions.

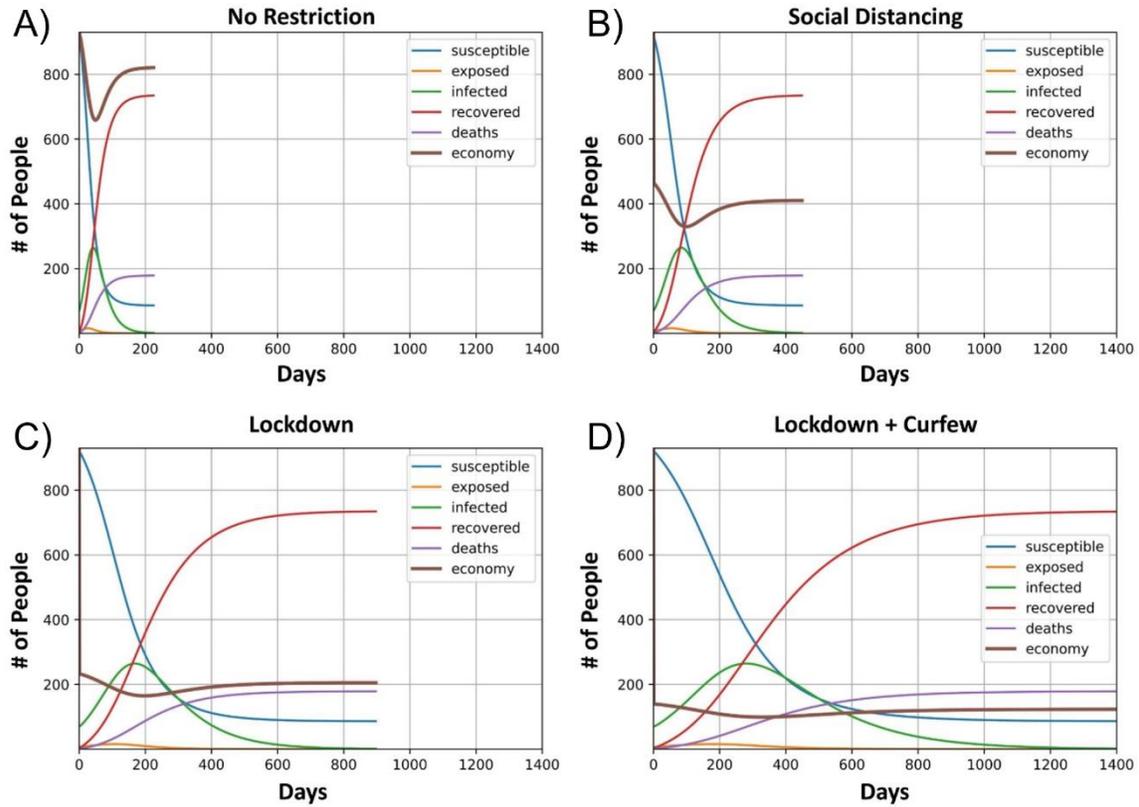

Figure 2. SEIRD models show susceptible (blue), exposed (orange), infected (green), recovered (red), and deceased (purple) populations along with economic status (brown) when different restrictive measures are used. The # of people in each population is seen in the y-axis and the number of days of the simulated pandemic is in the x-axis. Note that recovered and deceased populations are cumulative, whereas the other values are on a day-to-day basis. Additionally, note that on Day 0, the economic value was the total population (1000). However, on Day 1, due to the economic impact factor used to calculate the current economic value, each restriction seems to decrease the starting value of the economy, but the initial economic status is kept constant. The economy in our model is based on the number of people currently in the workforce (not infected or deceased), which is why the y-axis is the # of people for the economic line. The SEIRD model was solved using solve_ivp from SciPy and graphed using Matplotlib.

We trained two separate DRL agents with two separate goals (controlled by the reward function weights) on over 200 of the same pandemic scenarios, where they could implement any control strategy to find the optimal strategy for each pandemic (Figure 3). We found that after 200 of these cycles, the reward value began to plateau, indicating the agents converged to a near-optimal control strategy.

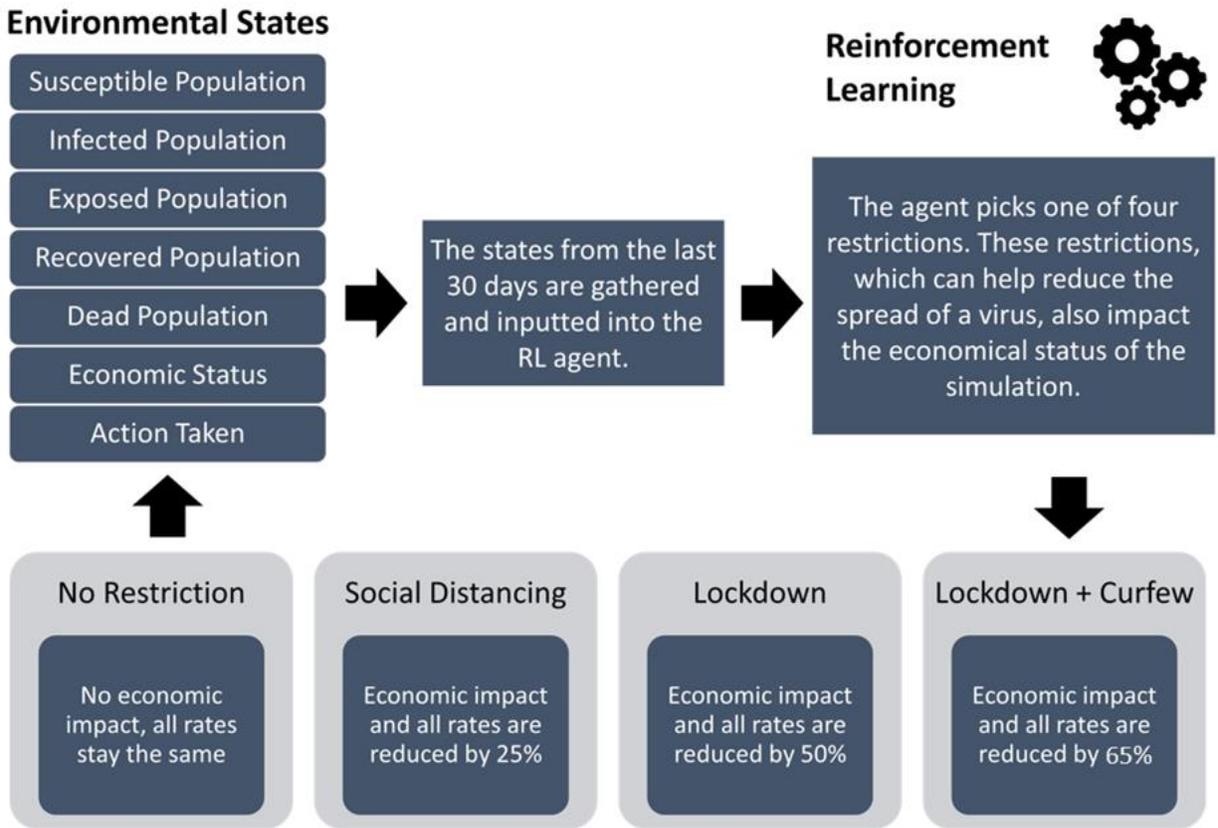

Figure 3. Diagram of a complete timestep during the agent training process. First, the current state of the SEIRD model (environment) was gathered. Then, the last 30 states were compiled and inputted into the DRL agent. The agent picked one of the four restrictions, which is applied to the SEIRD model rates. Finally, the next state of the model was calculated based on the updated rates and the previous state values. This whole process was simulated virtually using Python – specifically TensorFlow, SciPy, and Matplotlib. Economic impact and other SEIRD rates were reduced by fractions, mirroring real-world

pandemic data. This process was run for every single timestep (day) in a pandemic simulation for 200 training cycles.

The first agent aimed to balance the economy with pandemic effects (infection and deaths) whereas the second agent prioritized the economy over pandemic effects. Testing RL agents with multiple goals are vital for usage in the real world, since a different strategy may be used depending on the current economic state of the population. For example, a densely populated region might prioritize reducing infection, whereas a nation with pre-existing economic problems might have a bias toward protecting the economy.

After training, we tested the agents in the same environment and analyzed the policies and strategies they developed. The first agent balanced the economy and infection by beginning with a 10-day lockdown period, followed by cycles of 3 days of lockdown and 1 day of no restriction (Figure 4). Throughout this simulated pandemic, the reward continuously increased to a value of 1.2. While the reward value is arbitrary, since the model converged and plateaued at this reward, we can assume that this is the optimal reward value for when the infection rate is balanced with the economy. It should be noted that both the economy and infection rates stayed relatively constant throughout the first agent's best policy, indicating that neither had a bias that favored it.

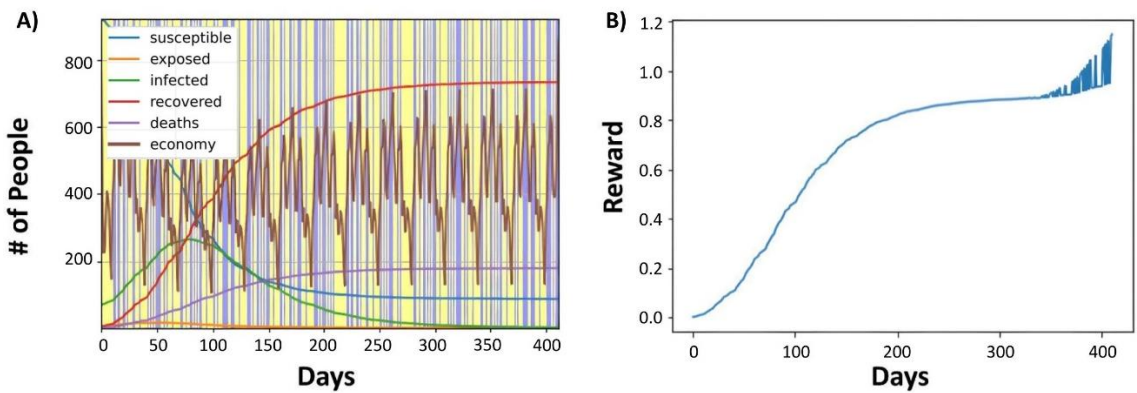

Figure 4. Control policy (yellow = lockdown, blue = no restriction), SEIRD results, and rewards from the unbiased agent with balanced weights between economy and infected cases (r = 12, s = 5). (A) SEIRD results of the first agent's decisions and their impact on the pandemic. The economy in our model is based on the number of people currently in the workforce (not infected or deceased), which is why the y-axis is the # of people for the economic line. The episode ends once the disease has completely been eradicated (*Infection* = 0). The restriction policy from the first agent consists of lockdowns and no restriction periods generated after 200 cycles of training the reinforcement learning agent. (B) Reward during this episode. After 200 days, the reward begins to flatten out.

The second agent prioritized economy over infection by following a 10-day lockdown and 20-day no-restriction cycle (Figure 5). Again, the reward continuously increased to a value of 1.6, confirming the second agent learned to maximize the reward more efficiently than the previous agent.

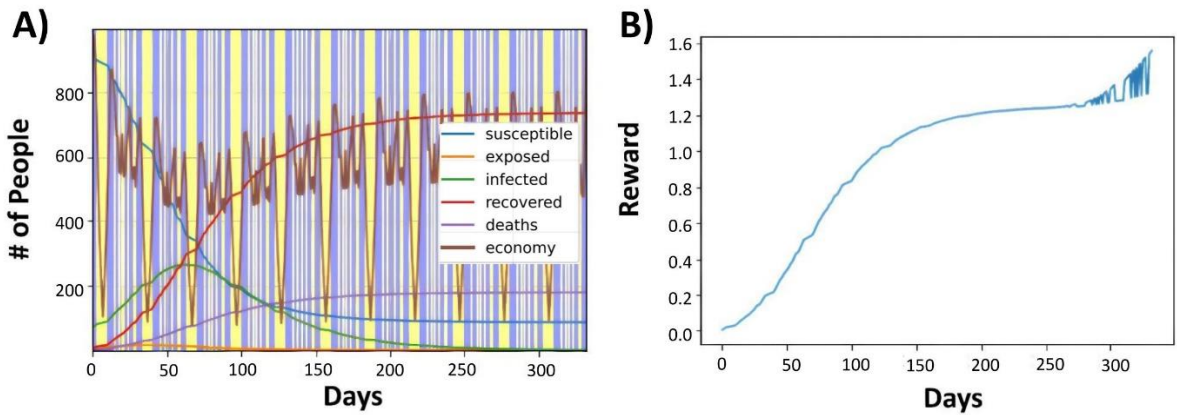

Figure 5. Control policy (yellow = lockdown, blue = no restriction), SEIRD results, and rewards from the economically-biased agent with weights preferring a greater economy over fewer cases (r = 10, s = 9). (A) SEIRD results of the second agent's decisions and their impact on the pandemic. In the background of the graph, yellow signifies a lockdown, whereas blue signifies no restriction. The economy in our model is based on the number of people currently in the workforce (not infected or deceased), which is why the y-

axis is the # of people for the economic line. The episode ends once the disease has completely been eradicated (*Infection* = 0). The restriction policy from the second agent consists of lockdowns and no restriction periods generated after 200 cycles of training the reinforcement learning agent. (B) Reward during this episode. After 150 days, the reward begins to flatten out.

From these results, we can see that the second agent provided a fluctuating, but on average much higher, economy with a slightly increased infection rate. Both agents had similar infectious results, with Agent 1 reaching quarter infection after 80 days and Agent 2 reaching quarter infection after 75 days (Figure 6). On the other hand, the economy was drastically different due to the different control policies used, with Agent 1 ending with an average economy of 40% normalcy and Agent 2 at 60% normalcy (Figure 7). It is also important to note that the economic trend with the second policy was increasing as the pandemic ends, indicating that it will likely continue to increase after the pandemic and reduce the risk of an economic crisis. Therefore, while the first strategy of balancing both infection and the economy were effective, the second strategy demonstrated that marginally higher infection rates significantly increase the economic activity in a population.

| Restriction | 25% Infected (days) | Economic Status (%) | Reward Peak |
|---|---|---|---|
| No Restriction | 50 | 60 | N/A |
| Social Distancing | 80 | 40 | N/A |
| Lockdown | 150 | 20 | N/A |
| Lockdown + Curfew | 300 | 15 | N/A |
| Balanced Agent | 80 | 40 | 1.2 |
| Economic Bias Agent | 75 | 60 | 1.6 |

Figure 6. Complete summary and comparison of infection rate, economic status, and reward peak while using each agent policy or single-restriction policy throughout. By reaching 25% infected later than with no restriction, the policies flatten the infection curve and allow for sufficient medical resources throughout the pandemic. Economic bias is seen since higher infection is sacrificed for a higher economy. Both perform similarly or better with the economy and infection than all current single-restriction strategies. Results were received by using the SEIRD model and two slightly different reinforcement learning agents. Policy results were generated after 200 cycles of training.

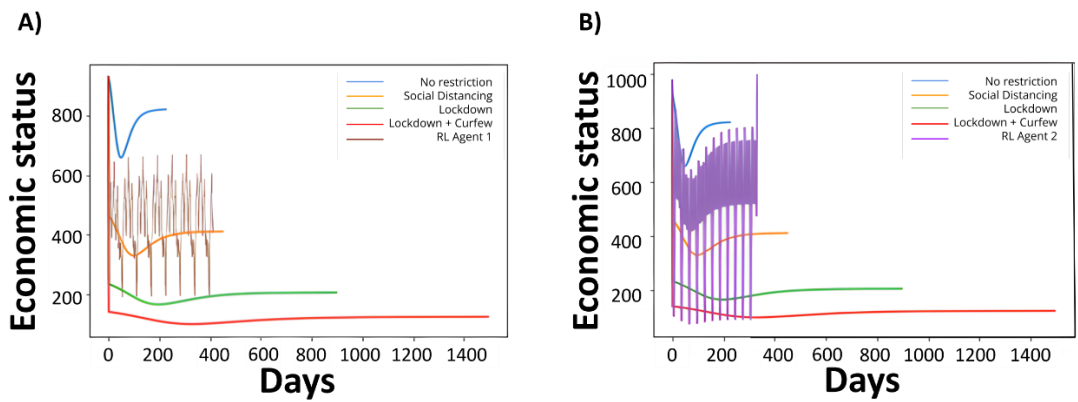

Figure 7. Economic comparisons between the balanced agent's policy (A), the economically-biased agent (B), and the single-restriction policies. The economy of the second, economically-biased, agent is significantly higher on average than the first, unbiased, agent, as expected. The economy is calculated by the number of people currently in the workforce. The unbiased agent has economic results similar to social distancing, whereas the economically-biased agent has economic results slightly lower than no restriction. Graphs were generated using the SEIRD model and Matplotlib.

We also found that the number of susceptible, exposed, recovered, and deceased individuals are equal at the end of both agents' simulations. This is expected as, due to the constant initial conditions of our SEIRD model, we will expect the same results each time since the restrictions are not directly affecting any of these factors. Rather, the restrictions affect only the infection rate and economic status directly, and, as a result of these restrictions, there is a downstream effect on the rest of the factors. More specifically, while the results for each of the factors stayed the same, in the unbiased first agent, it takes longer for it to reach the same number of susceptible, exposed, recovered, and deceased individuals since the rates of each are decreased due to the restrictions.

## 4. DISCUSSION

The first agent (which aimed to balance the economy and infection) had results that were similar to the single-action social distancing results, with both reaching 25% infected after 80 days and an average economic value of approximately 40% normalcy (Figure 6). This is not optimal as it did not offer an improved pandemic control strategy compared to any of the single-action results. However, even though this did not provide a novel strategy, it is supported by the high reward values which indicate that the agent did, to some extent, reach the goal of balancing economy and infection. As a result, this strategy could be utilized in a pandemic situation.

The second agent had an infection rate slightly higher than the first agent but an economy comparable to no restriction. This slight increase in infection rate is expected since this agent had an economic bias, causing it to sacrifice a higher infection rate for a significant economic increase. These results demonstrate that the implementation of the agent provided much better economic results than if only a single restriction had been used throughout the simulated pandemic.

These two agents have developed effective policies to control a pandemic. We can observe that the policies developed by each directly correlated to the changes in the initial reward function. As a result, this RL methodology could be applied to a wide range of pandemics to help understand how various restrictions affect both infection rates and the economy, and these factors can be balanced according to the needs of the country. The agent and the final policy also reacted sensitively to changes in the reward function, which confirms the experiment's goal-oriented functionalities.

Many different experiments can be conducted in the future to test this methodology and its applications. Future work may include testing a greater variety of reward function weights to find new and unique policies that could be implemented during pandemics. By adjusting the weights, we can test how the model would mitigate pandemics with varying control goals assigned by the population (ex. population biased towards protecting the economy). Additionally, we can test how the agent reacts to different initial values in the environment. This could also include modeling different types of economies by using more comprehensive and nuanced models. The current model has simplified the economy, the method by which infection spreads, and the effectiveness of each restriction. In the real world, the economy would be affected by other factors, infection spread could be dependent on geographical locations or age range, and the restrictions would not necessarily be followed and implemented by the whole population. We also could test the effectiveness of other pandemic strategies to control pandemics, such as masking and vaccinations. Finally, we could extend the SEIRD model to a multi-risk model which would enable us to test targeted lockdowns with an RL agent (7). With these extensions to the work conducted in this paper, we could create more accurate control strategies based on simulations that more closely resemble the real world.

This work offers a technique for pandemic control that can be applied by communities and governments to any pandemic, at any stage, and with any economic goals. As demonstrated by the results, the reinforcement learning agents effectively learned how to uniquely control a pandemic after only 200 training cycles, which on the average publicly available laptop can run in less than 2 days and with a supercomputer in less than 30 minutes. There is no significant evidence supporting the efficacy of the current strategies of only using one restriction, such as social distancing or lockdowns, and these restrictions tend to be put in place after disease spread has increased, rather than proactively. These measures are usually put into place without any specific plan and only aim to slow down disease spread to some extent until vaccines are available, which can take upward of six months (8). As a result, this training and generation methodology is well-suited for quickly controlling pandemics if sufficient data about the virus is available.

By following the same training techniques in any real-world scenario, the RL agent can provide an optimal strategy that considers infected cases and economic status. Additionally, any economic or infectious biases can be set in the reward function by adjusting the weights, and the agent will provide a policy that reacts accordingly, proving goal-oriented functionality. This offers a significant breakthrough in control policy generation, which currently does not take the economic factor into account. By applying this method, we can effectively develop novel, optimal strategies for control of any pandemic, protecting humanity from the effects of future pandemics.

## 5. MATERIALS AND METHODS

*Epidemic simulation*

We utilized a SEIRD compartmental model, which illustrates the state (susceptible, exposed, infected, recovered, and deceased) of a population throughout a pandemic and uses ordinary differential equations (ODEs) (9). Once initial parameters were set, the model was able to simulate the progression of individuals through the five different stages of infection, which is modeled (Figure 1). The ODEs that make up the SEIRD model can be solved by substituting the following initial conditions used by Quwsar et. al to simulate a pandemic: N = total population = 1000 people, β = transmission rate = 0.12 contacts/day, σ = incubation period = 1 day, γ = recovery rate = 1/27 recoveries/day, μ = death rate = 0.009 mortalities/day, and I = infected population = 7% of the total population (13):

$$S = Susceptible\ population: \frac{dS}{dt} = -\frac{\beta IS}{N}$$

$$E = Exposed\ population: \frac{dE}{dt} = \frac{\beta IS}{N} - \sigma E$$

$$I = Infected\ population: \frac{dI}{dt} = \sigma E - (\gamma + \mu)I$$

$$R = Recovered\ population: \frac{dR}{dt} = \gamma I$$

$$D = Deceased\ population: \frac{dD}{dt} = \mu I$$

To solve the ODEs, which are initial value problems, we utilized a SciPy module known as solve_ivp and the Runge-Kutta method (10, 11). Using this process, we solved the equations at every timestep (day) in the pandemic simulation and retrieved the state information for each day.

*Reinforcement learning*

To enable an RL agent to balance infectious cases and the economy during a pandemic, we first assigned a numerical value to the economic status on any given day alongside the epidemiological model. We completed this task by utilizing a custom economic equation:

$$\frac{dE_{econ}}{dt} = (S + E + R) \times (1 - \zeta)$$

This equation uses the portion of the population that can contribute economically by working (Susceptible, Exposed, and Recovered individuals) scaled by a factor, $\zeta$, that represents the economic impact of the current restriction (no restriction: 0, social distancing: 0.25, lockdown: 0.5, lockdown + curfew: 0.65). These values are based on estimates from past pandemics (2). We utilized this value throughout the pandemic simulation to provide a general understanding of the economic status and also test the impact of restrictions such as social distancing and lockdowns.

In RL, reward functions are used to reward certain actions taken by an agent and penalize others to train the agent and formulate the optimal policy (12). By using the correct reward function (that is, one that takes into account all important factors that the agent must take into consideration), we were able to teach our agent to balance pandemic effects, such as infectious cases and deaths, with the economy. In our work, we utilized a custom reward function proposed by Quwsar, et al., where R = reward, E = economic status, r = weight of infected cases vs. economy = 10, s = weight of deaths = 7, A = % currently infected, and D = % deceased,

$$R(s) = E \times e^{-r \times A} - s \times D$$

which has proven results in teaching a reinforcement learning agent accurately (13). This reward function took into consideration infectious cases, deaths, and economic status, allowing it to provide rewards for every action taken by the agent. It also included adjustable weights for each factor, *r,* and *s*, that can be set to balance the economy and pandemic effects or prioritize one over the other. By increasing *r,* the agent prioritized infectious cases increasingly. In this work, we used two distinct values for the weights to train two separate agents and test goal-oriented functionalities. Specifically, the first agent aimed to balance the economy and the pandemic effects ($r = 12$, $s = 5$) as used by Quwsar et al. in a balanced simulation (13). On the other hand, the second agent prioritized economic status over pandemic effects ($r = 10$, $s = 9$), which is seen by decreasing the r-value (therefore decreasing the priority of infected cases and increasing the priority of the economy) and increasing s value (increasing the priority of deaths, which is essential so that the agent does not completely disregard the disease in favor of maintaining the economy).

In DRL, an agent first receives the current state of the environment, *S*. Then, this state is inputted into a deep neural network, which outputs an action, *a,* that the algorithm believes will provide the highest reward (14). Finally, this action interacts with the environment and provides a new state, *s'*. By utilizing a reward function to calculate the reward at *s'*, the agent learns whether or not its action was positive and updates its neural network accordingly. By looping this process for *n* training cycles, the agent formulates an optimal policy that will provide high rewards consistently.

We used a Deep Double Q-Network (DDQN) method to train our agent (6). A vanilla Q-learning algorithm consists of a Q-table that takes the state and action taken in an environment and assigns a Q-value to it, which is the reward (15). Based on this table, the algorithm can make the best decision in every state by choosing the action with the highest reward. In DDQN, on the other hand, instead of requiring every state and action pairing to make the best decision, it can use two neural networks to make decisions: the Deep Q-network and a target network. The Deep Q-network picks the action with the highest reward for the next state, whereas the target network predicts the reward if the action is taken. Then, the Deep Q-network is updated based on the prediction from the target network.

We utilized a Bidirectional Long Short-Term Memory (Bi-LSTM) neural network architecture with a 30-day memory as the deep neural network for our reinforcement learning agent (Figure 8) (14). This 30-day memory interval was found to be optimal by Quwsar, et al. to allow the agent to learn both the immediate and delayed effects of implementing each restrictive measure (13). A Bi-LSTM layer allowed the agent to make forward and backward connections between the time-series data provided by a pandemic. By enabling the agent to make these connections over a longer period, such as 30 days, it was able to learn how certain actions have not only immediate impacts but also delayed effects. By using the information from the past 30 days along with prior learning, the agent picked from one of four actions: no restriction, social distancing, lockdown, or lockdown and curfew. We then attached specific pandemic advantages and economic disadvantages to each restriction based on real-world data (Figure 3, 16 - 19). Additionally, we used the epsilon-greedy policy, where ε = 1 and gradually decays as *ε = max (1- ε /125, 0.01)*, allowing for sufficient exploration of the environment (20). Finally, mean squared error (MSE) was used to calculate the loss of the DDQN, as it compared the predicted reward from the network with the actual reward (21).

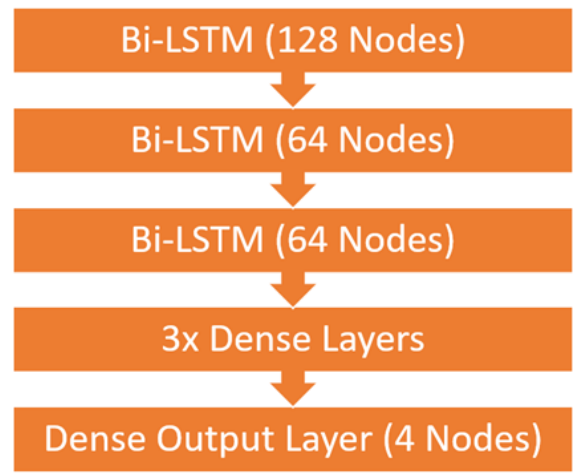

Figure 8. RL agent neural network model architecture. We utilized three Bi-LSTM layers to make forward and backward connections between input data. Then, dense layers were used to output one of four actions. The input to the model is of shape (1,30,7). Individual states have seven pieces of information and we inputted the last 30 days of states. The model was created using TensorFlow,

specifically Keras. In this work, two agents with identical network structures were trained with modified reward functions.

We repeated this training over 200 cycles, where one cycle is a complete pandemic simulation from the initial outbreak until complete disease eradication. As a result, different simulations have different lengths, because, based on the actions taken during the simulation, it may take longer or shorter for the disease to become completely eradicated from the population, which is when the simulation stops.

This complete implementation was conducted using Python, Keras, TensorFlow, and NumPy (22-24). Matplotlib was utilized for visual models (25). A GitHub page is available to access the code used for the simulation and reinforcement learning implementation (26).

## 6. ACKNOWLEDGEMENTS

We would like to thank Dr. Yelena Naumova for her support and encouragement throughout our project.

## 7. REFERENCES

1. Watkins, John. "Preventing a covid-19 pandemic." *BMJ (Clinical research ed.),* vol. 368, no. 810. 28 Feb. 2020, doi:10.1136/bmj.m810.

2. Pak, Anton, et al. "Economic Consequences of the COVID-19 Outbreak: the Need for Epidemic Preparedness." *Frontiers in public health*, vol. 8, no. 241, 29 May 2020, doi:10.3389/fpubh.2020.00241.

3. Van Hasselt, Hado, et al. "Deep Reinforcement Learning with Double Q-learning." *Machine Learning*, 8 Dec. 2015, doi:10.48550/arXiv.1509.06461.

4. Mnih, V. et al. "Human-level control through deep reinforcement learning." *Nature*, vol. 518, no. 7540, 25 Feb. 2015, doi: 10.1038/nature14236.

5. Zhang, Yi, et al. "Human-like Autonomous Vehicle Speed Control by Deep Reinforcement Learning with Double Q-Learning." *2018 IEEE Intelligent Vehicles Symposium (IV)*, 28 June 2018, doi:10.1109/ivs.2018.8500630.

6. Moreno-Vera, Felipe, et al. "Performing Deep Recurrent Double Q-Learning for Atari Games," *2019 IEEE Latin American Conference on Computational Intelligence (LA-CCI),* 15 Nov. 2019, doi:10.1109/LA-CCI47412.2019.9036763.

7. Acemoglu, Daron, et al. "Optimal Targeted Lockdowns in a Multigroup SIR Model." *American Economic Review: Insights*, vol. 3, no. 4, 1 Dec. 2021, pp. 487–502, 10.1257/aeri.20200590.

8. Markel, Howard, et al. "Nonpharmaceutical interventions implemented by US cities during the 1918-1919 influenza pandemic." *JAMA,* Aug. 2007, doi:10.1001/jama.298.6.644

9. Biswas, M. H. A., et al. "A SEIR Model for Control of Infectious Diseases with Constraints." *Mathematical Biosciences and Engineering*, vol. 11, no. 4, Mar. 2014, pp. 761–784, doi:10.3934/mbe.2014.11.761.

10. Virtanen, Pauli, et al. "SciPy 1.0: Fundamental Algorithms for Scientific Computing in Python." *Nature Methods*, vol. 17, no. 3, 3 Feb. 2020, pp. 261–272, doi:10.1038/s41592-019-0686-2.

11. Cash, J. R., and Alan H. Karp. "A Variable Order Runge-Kutta Method for Initial Value Problems with Rapidly Varying Right-Hand Sides." *ACM Transactions on Mathematical Software (TOMS)*, vol. 16, no. 3, Sept. 1990, pp. 201–222, doi:10.1145/79505.79507.

12. Hu, Zijian, et al. "A Dynamic Adjusting Reward Function Method for Deep Reinforcement Learning with Adjustable Parameters." *Mathematical Problems in Engineering*, vol. 2019, 23 Nov. 2019, pp. 1–10, doi:10.1155/2019/7619483.

13. Ohi, Abu Quwsar, et al. "Exploring Optimal Control of Epidemic Spread Using Reinforcement Learning." *Scientific Reports*, vol. 10, no. 1, Dec. 2020, doi:10.1038/s41598-020-79147-8.

14. Graves, Alex, et al. "Hybrid Speech Recognition with Deep Bidirectional LSTM." *2013 IEEE Workshop on Automatic Speech Recognition and Understanding*, 8 Dec. 2013, doi: 10.1109/ASRU.2013.6707742.

15. O'Donoghue, Brendan et al. "Combining policy gradient and Q-learning." *Machine* Learning, 7 Apr. 2017, doi:10.48550/arXiv.1611.01626.

16. Gibney, Elizabeth. "Whose Coronavirus Strategy Worked Best? Scientists Hunt Most Effective Policies." *Nature*, 27 Apr. 2020, doi:10.1038/d41586-020-01248-1.

17. Oraby, Tamer, et al. "Modeling the Effect of Lockdown Timing as a COVID-19 Control Measure in Countries with Differing Social Contacts." *Scientific Reports*, vol. 11, no. 1, 8 Feb. 2021, doi:10.1038/s41598-021-82873-2.

18. VoPham, Trang, et al. "Effect of Social Distancing on COVID-19 Incidence and Mortality in the US." *PubMed*, 12 Jun. 2020, doi:10.1101/2020.06.10.20127589.

19. Andronico, Alessio, et al. "Evaluating the Impact of Curfews and Other Measures on SARS-CoV-2 Transmission in French Guiana." *Nature Communications*, vol. 12, no. 1, 12 Mar. 2021, doi:10.1038/s41467-021-21944-4.

20. Wunder, Michael, et al. "Classes of Multiagent Q-Learning Dynamics with -Greedy Exploration." *ICML 2010*, 16 Jul. 2019, doi: 10.1145/1553374.1553422.

21. Wallach, D., and B. Goffinet. "Mean Squared Error of Prediction as a Criterion for Evaluating and Comparing System Models." *Ecological Modelling*, vol. 44, no. 3-4, Jan. 1989, pp. 299–306, doi:10.1016/0304-3800(89)90035-5.

22. Raschka, Sebastian, et al. "Machine Learning in Python: Main Developments and Technology Trends in Data Science, Machine Learning, and Artificial Intelligence." *Information*, vol. 11, no. 4, 4 Apr. 2020, p. 193, doi:10.3390/info11040193.

23. Abhishek Nandy, and Manisha Biswas. *Reinforcement Learning: With Open AI, TensorFlow and Keras Using Python*. New York Apress, 2018.

24. van der Walt, Stéfan, et al. "The NumPy Array: A Structure for Efficient Numerical Computation." *Computing in Science & Engineering*, vol. 13, no. 2, Mar. 2011, pp. 22–30, doi:10.1109/mcse.2011.37.

25. Hunter, John D. "Matplotlib: A 2D Graphics Environment." *Computing in Science & Engineering*, vol. 9, no. 3, 2007, pp. 90–95, doi:10.1109/mcse.2007.55.

26. GitHub page for code used for the simulation and reinforcement learning implementation: https://ishirraov.github.io/